\documentclass{article}

\usepackage[dblblindworkshop,final]{neurips_2025}

\workshoptitle{MATH-AI}

\usepackage[utf8]{inputenc} % allow utf-8 input
\usepackage[T1]{fontenc}    % use 8-bit T1 fonts
\usepackage{hyperref}       % hyperlinks
\usepackage{url}            % simple URL typesetting
\usepackage{booktabs}       % professional-quality tables
\usepackage{amsfonts}       % blackboard math symbols
\usepackage{nicefrac}       % compact symbols for 1/2, etc.
\usepackage{microtype}      % microtypography
\usepackage[table]{xcolor} % color for tables
\usepackage{xcolor}         % colors
\usepackage{graphicx}       % For including images
\usepackage{enumitem}       % For custom lists
\usepackage{amsmath}
\usepackage{wrapfig,multirow,colortbl}
\usepackage[misc]{ifsym}

\title{Bridging Vision, Language, and Mathematics: Pictographic Character Reconstruction with Bézier Curves}

\author{\textbf{Zihao Wan}\textsuperscript{1}\footnotemark[1], \textbf{Pau Tong Lin Xu}\textsuperscript{1}\footnotemark[1], 
\textbf{Fuwen Luo}\textsuperscript{1}, \textbf{Ziyue Wang}\textsuperscript{1}, \\ \textbf{Peng Li}\textsuperscript{2\ \Letter}, \textbf{Yang Liu}\textsuperscript{1,2\ \Letter}\\ 
\textsuperscript{1}Dept. of Comp. Sci. \& Tech., Institute for AI, Tsinghua University, Beijing, China \\
\textsuperscript{2}Institute for AI Industry Research (AIR), Tsinghua University, Beijing, China \\
}

\begin{document}

\maketitle

\renewcommand{\thefootnote}{\fnsymbol{footnote}} 
    \footnotetext[1]{Equal contribution, \textsuperscript{\Letter} Corresponding author}
\renewcommand{\thefootnote}{\arabic{footnote}}

\begin{abstract}
  While Vision-language Models (VLMs) have demonstrated strong semantic capabilities, their ability to interpret the underlying geometric structure of visual information is less explored. Pictographic characters, which combine visual form with symbolic structure, provide an ideal test case for this capability. We formulate this visual recognition challenge in the mathematical domain, where each character is represented by an executable program of geometric primitives. This is framed as a program synthesis task, training a VLM to decompile raster images into programs composed of Bézier curves. Our model, acting as a ``visual decompiler'', demonstrates performance superior to strong zero-shot baselines, including GPT-4o. The most significant finding is that when trained solely on modern Chinese characters, the model is able to reconstruct ancient Oracle Bone Script in a zero-shot context. This generalization provides strong evidence that the model acquires an abstract and transferable geometric grammar, moving beyond pixel-level pattern recognition to a more structured form of visual understanding.
\end{abstract}

% \vspace{-3pt}

\begin{figure}[h]
  \centering
  \vspace{-9pt}
  \includegraphics[width=\linewidth]{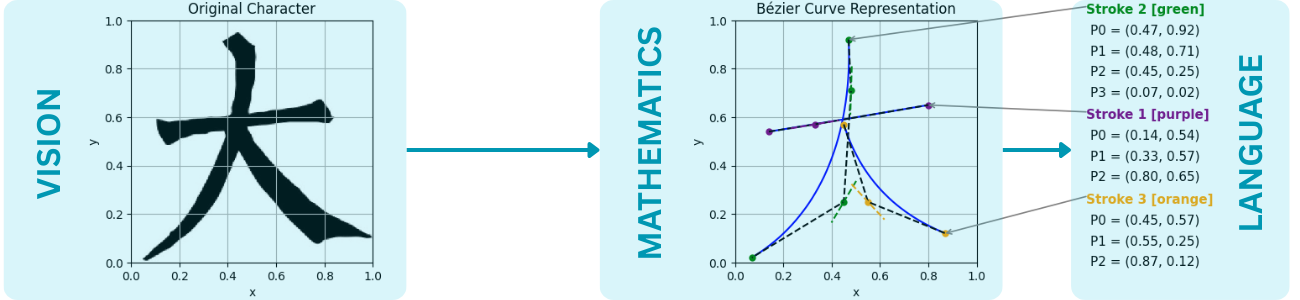} 
  \vspace{-18pt}
  \caption{\textbf{Mathematics bridging vision and language.} A pictographic character can be represented using Bézier curves, each of which can then be written as a list of control points in text form.}
  \label{fig:vision-math-text}
% \vspace{-12pt}
\end{figure}

\section{Introduction}

% 这里就要说出我们的motivation，因为Vision-language Models (VLMs) are promising xxxx，所以对于文字以及图画的识别（recognize）也有一定能力。象形文字是从图演化来的，那么这让我们思考，vlm到图文识别和理解能力是否能泛化到对象形文字文字的recostruct上。我们发现不太行，因为在这个任务上他们有xxxx问题（或者由于欠缺什么能力，导致他们学不好）
Vision-language Models (VLMs) have demonstrated strong capabilities in mapping images to high-level semantic descriptions~\citep{bai2025qwen25vltechnicalreport,openai2024gpt4ocard}. However, a more fundamental aspect of visual understanding — the ability to interpret and reconstruct the underlying structure of visual forms — has been less explored~\citep{thrush2022winogroundprobingvisionlanguage}. Pictographic characters, which combine pictorial representation with a structured symbolic system, present an ideal test case for this capability, motivating our core question: \textit{can VLMs' understanding generalize from semantic recognition to programmatic reconstruction?} We find that current models, including GPT-4o, are limited in this capacity (\autoref{fig:case-study}). This difficulty suggests that current VLMs rely on surface-level correlations between pixel patterns and linguistic tokens, rather than parsing visual forms into geometric programs.

%第一句话：考虑到文字由线条组成，我们用mathematical curves（Bézier curves）去解决上面提到的问题。（注意Bézier需要引用，参考其他用Bézier curves的论文里引用）
%第二/三句：具体强调了什么/给模型引入了什么能力，让模型能够reconstruct。
%第四句：简单说方法We train a VLM to act as a visual decompiler: given an image, it outputs a sequence of Bézier curves. This xxxxx
To address these limitations, we frame character reconstruction as a program synthesis problem, with the target output being a sequence of mathematical primitives — Bézier curves. This representation compels the model to move beyond pixel-level statistics and learn the geometric properties that define a character's form. Unlike localization methods such as bounding boxes~\citep{Girshick_2014_CVPR, NIPS2015_14bfa6bb}, a Bézier sequence provides a fine-grained, executable representation of an object's continuous vector shape. We train a VLM to act as a ``visual decompiler'', inferring the underlying generative program from its rendered, pixel-based output (\autoref{fig:vision-math-text}). This allows for substantial generalization: a model trained only on modern Chinese characters can reconstruct Oracle Bone Script in a zero-shot setting, suggesting an internalized ``geometric grammar'' — the compositional rules governing the arrangement, orientation, and form of the geometric primitives that constitute a character's structure.

% para 1: 数学的作用，用数学把文字和像素桥接起来
% gpt等直接做不行（导向fig. 1）

% 说明这个任务没有其他人做过

%三个contributions
%1.我们提出用贝塞尔交模型写字（不一定直接这么说，但是表达这个意思），说明这个任务（是任务，不是方法，不是说novel framework）没有其他人做过
%2.A novel framewor xxx
%3.Empirical evidence that our models outperform xxx

Our contributions are as follows:
\vspace{-9pt}
% \begin{itemize}[leftmargin=*]
\begin{itemize}[left=0.4cm, itemsep=2pt, parsep=0pt]
    \item We introduce and formalize the task of programmatic character reconstruction, where a model decompiles a raster image into an executable program of Bézier curves. To our knowledge, this is the first work to define character understanding as one of direct geometric program synthesis.
    \item We propose a framework that trains a VLM as a ``visual decompiler'', directly translating raster images into Bézier curve sequences with the spatial aid of an explicit visual coordinate system.
    \item We provide evidence that our model substantially outperforms strong zero-shot baselines such as GPT-4o. Crucially, it achieves zero-shot cross-script generalization from modern to ancient characters (\autoref{tab:main-results}), verifying its ability to learn an abstract and transferable geometric representation.
\end{itemize}
% \vspace{-12pt}

% \vspace{-3pt}
\section{Methodology}
% \vspace{-6pt}

% para 1: 简单说：我们的方法主要包括提取贝塞尔曲线和构建训练数据
% para 2: 新开一段，讲提取贝塞尔曲线（构造图到贝塞尔的关系）
% para 3: 新开一段，讲构建训练数据（贝塞尔曲线的文字形式，和坐标轴，这两个分别开两个subsection）
% method 写长一点，现在空间剩出来比较多，不单说我们的方法是什么，以及要说为什么好，比如构建贝塞尔，就是好在我们是通过规则化的方法去造，无需公开数据集，还可以佐以一个case说明这样，包括我们怎么拆的笔画，就是造出来的数据跟汉字本身的笔画有多相似之类的

In this section, we introduce how we construct paired ``image-Bézier curve'' training data and train VLM. We first introduce the data construction process of deriving Bézier curves from images of pictographic characters in Section~\ref{sec:curve_extraction}, then introduce our training paradigm in Section~\ref{sec:training_paradigm}.

% Our methodology reframes character understanding as a geometric reconstruction task through a two-stage process: we first extract a structured geometric representation (Bézier curves) from a raster image, then serialize this representation into a textual program suitable for training a Vision-Language Model. A detailed mathematical background on Bézier curves is provided in \autoref{sec:appendix_bezier}.

% \vspace{-6pt}
\subsection{Bézier Curves Extraction}\label{sec:curve_extraction}
% \vspace{-3pt}

To the best of our knowledge, few previous works have attempted to automatically construct mathematical representations, such as Bézier curves, from images of pictographic characters. Manual annotations, on the other hand, are costly and limited in scale, posing challenges for models to learn such representations and limiting their abilities to perform programmatic reconstruction. Thus, we develop an automated rule-based pipeline that converts character images to Bézier curve sequences.
% To generate training data at scale, we developed an automated pipeline that converts character images into Bézier curve sequences.
This approach allows us to create a large-scale ``image-Bézier curve'' dataset from any font file efficiently without requiring manual annotations. The pipeline operates in four steps:
\begin{enumerate}[left=0.4cm, itemsep=2pt, parsep=0pt]
\vspace{-8pt}
    \item \textbf{Preprocessing}: Input images are binarized to isolate the glyph.
    \item \textbf{Skeletonization}: The glyph is skeletonized to a one-pixel-wide centerline, which preserves its topology.
    \item \textbf{Graph Segmentation}: An 8-connected pixel graph~\citep{beziergen} is built from the skeleton and segmented into paths at junctions and endpoints.
    \item \textbf{Curve Fitting}: Each path is simplified using the Ramer-Douglas-Peucker algorithm~\citep{douglas1973algorithms}, and adjacent segments are iteratively merged based on proximity and alignment to form smooth strokes.
\end{enumerate}
\vspace{-8pt}
This process yields a clean list of Bézier curves that faithfully represents the character's strokes. Further details are provided in \autoref{sec:python_libraries}.

% \vspace{-6pt}
\subsection{Training Paradigm}\label{sec:training_paradigm}
% \vspace{-3pt}

% 为了让VLM理解贝塞尔曲线，我们首先把贝塞尔曲线用文字表示
A primary challenge for a VLM to generate geometric programs is bridging the its textual output with the visual and spatial nature of target shapes. To address this, we first serialize the geometric information. We represent each glyph as a sequence of strokes encapsulated within ``<bezierseq>'' tags. Each stroke is defined by its control points inside ``<bezier>'' tags, e.g., \texttt{\small <bezier>(x1 y1)...</bezier>}. All coordinate values are normalized to a $[0, 1]$ range.  This serialization transforms the visual reconstruction task into a program synthesis problem, compelling the model to learn a structured, symbolic representation of geometry rather than relying solely on pixel-level correlations.
% 然而，仅用文字表示是不够的，模型因为空间理解能力不足...，重构效果不（引用实验）好，因此我们引入了...

However, serialization alone proves insufficient, as our analysis confirms that VLMs struggle to ground abstract coordinates to absolute positions within an image (see \autoref{tab:axis-ablation}). To overcome this critical spatial reasoning deficit, we augment the input image by overlaying a normalized Cartesian coordinate system on the glyph. This visual aid, consisting of labeled x- and y-axes, provides explicit spatial context. This visual framework provides an explicit spatial anchor, enabling the model to directly map visual features to their precise numerical locations.

\vspace{-6pt}
\section{Experiments and Results}
\vspace{-3pt}

\subsection{Experimental Settings}
% \vspace{-6pt}
\paragraph{Models and Frameworks} We use the Qwen2.5-VL-7B model~\citep{bai2025qwen25vltechnicalreport} as our backbone, leveraging its strong vision-language capabilities. All experiments are conducted using the MS-Swift~\citep{zhao2024swiftascalablelightweightinfrastructure} training framework. We compare our method against strong zero-shot baselines such as GPT-4o~\citep{openai2024gpt4ocard} and the base Qwen2.5-VL-7B. More details on our baseline configurations are provided in \autoref{sec:appendix_exp}.

% \vspace{-9pt}
\paragraph{Training Data} Our primary training dataset consists of 2,000 common Chinese characters. We use six font variations to create a dataset of approximately 12,000 samples for our main Supervised Fine-tuning (SFT) experiments. \footnote{\scriptsize Six fonts variations (Microsoft YaHei, SimHei, SimKai, SimLi, SimYou, SimSun), are standard system fonts included with Microsoft Windows.}

% \vspace{-9pt}
\paragraph{Evaluation Data} We evaluate our models on three challenging, unseen datasets:
\vspace{-6pt}
% \begin{itemize}[leftmargin=*]
\begin{itemize}[left=0.2cm, itemsep=2pt, parsep=0pt]
    \item \textbf{Chinese STD:} 1,000 unseen Chinese characters in a standard font (Source Han Sans SC Normal).
    \item \textbf{Chinese Stylistic:} 1,000 unseen Chinese characters in a stylistic, calligraphy-like font (JinNianYeYaoJiaYouYa) \footnote{\scriptsize The modern Chinese fonts used for evaluation, Source Han Sans SC and JinNianYeYaoJiaYouYa, were sourced from the public font website \url{https://www.fonts.net.cn/}.}, which resembles handwritten style and helps assess the model’s ability to capture expressive and diverse stroke patterns.
    \item \textbf{OBS:} 1,000 unseen Oracle Bone Script characters from the HWOBC~\citep{HWOBC} dataset. This serves as our primary test set for cross-script generalization.
\end{itemize}
\vspace{-6pt}

\begin{table}
  \caption{Main results comparing our trained model against zero-shot baselines. We report on three sub-metrics (D: Distance, A: Angle, L: Length) and the final geometric reconstruction score (G). Our model demonstrates superior performance and remarkable cross-script generalization.}
  \label{tab:main-results}
  \vspace{-6pt}
  \centering
  \scriptsize
  % Further reduce column spacing for a more compact table
  % \setlength{\tabcolsep}{3.5pt}
  \resizebox{\textwidth}{!}{
  \begin{tabular}{l c@{\hspace{0.2cm}}c@{\hspace{0.2cm}}c@{\hspace{0.3cm}}c c@{\hspace{0.2cm}}c@{\hspace{0.2cm}}c@{\hspace{0.3cm}}c c@{\hspace{0.2cm}}c@{\hspace{0.2cm}}c@{\hspace{0.3cm}}c}
    \toprule
     \multirow{2}{*}[-0.9ex]{\textbf{\textbf{Model / Method}}} & \multicolumn{4}{c}{\textbf{Chinese STD (STD-zh)}} & \multicolumn{4}{c}{\textbf{Chinese Stylistic (Stylistic-zh)}} & \multicolumn{4}{c}{\textbf{OBS}} \\
    \cmidrule(lr){2-5} \cmidrule(lr){6-9} \cmidrule(lr){10-13}
    & D$\uparrow$ & A$\uparrow$ & L$\uparrow$ & \cellcolor{blue!8} G$\uparrow$ & D$\uparrow$ & A$\uparrow$ & L$\uparrow$ & \cellcolor{blue!8} G$\uparrow$ & D$\uparrow$ & A$\uparrow$ & L$\uparrow$ & \cellcolor{blue!8} G$\uparrow$ \\
    \midrule
    Qwen2.5-VL-7B (Zero-Shot) & 0.019 & 0.012 & 0.029 & \cellcolor{blue!8} 0.157 & 0.023 & 0.019 & 0.037 & \cellcolor{blue!8} 0.177 & 0.033 & 0.024 & 0.049 & \cellcolor{blue!8} 0.182 \\
    GPT-4o (Zero-Shot) & 0.139 & 0.119 & 0.212 & \cellcolor{blue!8} 0.431 & 0.201 & 0.149 & 0.288 & \cellcolor{blue!8} 0.467 & 0.163 & 0.105 & 0.232 & \cellcolor{blue!8} 0.409 \\
    \midrule
    Ours (Qwen2.5-VL-7B + Warmup + RL) & 0.508 & 0.454 & 0.654 & \cellcolor{blue!8} 0.678 & 0.561 & 0.463 & 0.687 & \cellcolor{blue!8} 0.663 & 0.359 & 0.292 & \textbf{0.478} & \cellcolor{blue!8} 0.548 \\
    \textbf{Ours (Qwen2.5-VL-7B + SFT)} & \textbf{0.770} & \textbf{0.686} & \textbf{0.854} & \cellcolor{blue!8} \textbf{0.821} & \textbf{0.603} & \textbf{0.509} & \textbf{0.696} & \cellcolor{blue!8} \textbf{0.723} & \textbf{0.362} & \textbf{0.293} & 0.463 & \cellcolor{blue!8} \textbf{0.568} \\
    \bottomrule
  \end{tabular}
  }
  \vspace{-15pt}
\end{table}

% \vspace{-9pt}
\paragraph{Experimental Details} Our primary training method is Supervised Fine-Tuning (SFT)~\citep{NEURIPS2022_b1efde53}, where the main objective is to teach the VLM to function as a ``visual decompiler'' that translates character images into their corresponding Bézier curve programs. The model is trained on the axis-enhanced images and their ground-truth programs using a standard next-token prediction objective. For comparison, we also experiment with Reinforcement Learning (RL). This involves initializing a model from an SFT-warmed-up checkpoint and continuing training with a geometric reward signal using the GRPO algorithm~\citep{shao2024deepseekmathpushinglimitsmathematical}. This setup enables a direct comparison between purely supervised learning and reward-driven exploration. Further training details are available in \autoref{sec:appendix_exp}.

% \vspace{-9pt}
\paragraph{Evaluation Metrics}
We evaluate reconstruction quality using a comprehensive \textit{Geometric Score} (G) as our primary metric, which also serves as the reward signal in our RL experiments. To provide a more granular analysis, we additionally report three sub-scores derived from the same matching framework: \textit{Distance}, \textit{Angle}, and \textit{Length}. The primary \textit{Geometric Score} calculation holistically compares the generated and ground-truth Bézier curve sequences by finding an optimal one-to-one stroke matching based on a composite similarity function. All final scores are normalized into the range $[0, 1]$, where 1 signifies a perfect match. A detailed breakdown is provided in \autoref{sec:appendix_eval}.

% \vspace{-9pt}
\subsection{Main Results}
% \vspace{-6pt}
Our findings, detailed in \autoref{tab:main-results}, confirm the effectiveness of our method across the overall reconstruction score and all three sub-metrics. Our model achieves the best results across all investigated datasets, demonstrating robust generalization to unseen standard and stylistic fonts.

\vspace{-3pt}
Most strikingly, the model exhibits remarkable cross-script generalization. On the zero-shot Oracle Bone Script (OBS) evaluation, our SFT model scores 0.568, a \textbf{15.9\%} increment over the powerful GPT-4o baseline. This result strongly supports our hypothesis that by learning programmatic construction, the model acquires a transferable geometric grammar rather than memorizing pixel patterns. To validate our Geometric Score's effectiveness, we conducted a human evaluation comparing our SFT model against GPT-4o on 150 samples in \autoref{sec:appendix_human_eval}. The results, showing a 91.6\% average win rate for our model, confirm that our quantitative metric aligns well with human qualitative judgment.

% \vspace{-6pt}
\section{Analysis}
% \vspace{-6pt}

\subsection{Case Study: Visualizing Reconstruction}

\begin{figure}
  \centering
  \includegraphics[width=\linewidth]{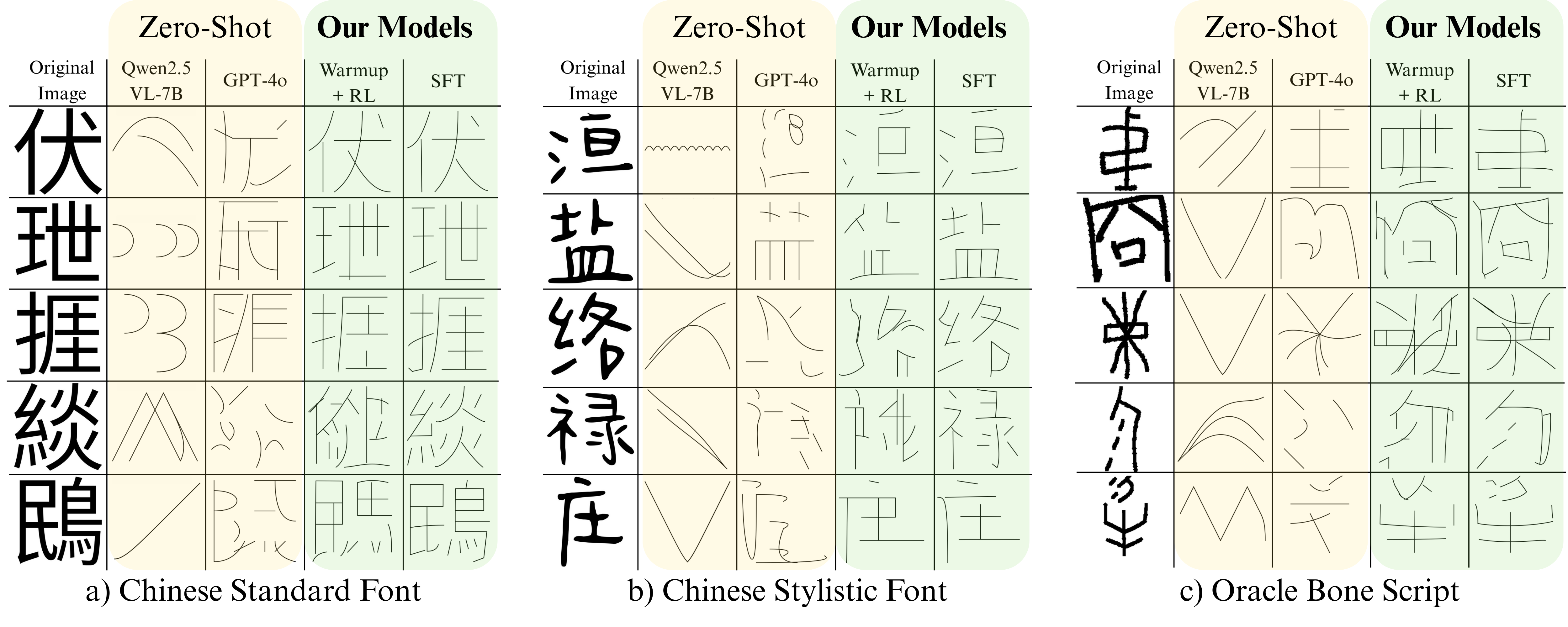} 
  \vspace{-18pt}
  \caption{\textbf{Qualitative comparison of reconstructions across models.} \textbf{a, b, c} show reconstruction cases of Chinese standard font, Chinese stylistic font, and Oracle Bone Script, respectively.}
  \label{fig:case-study}
  \vspace{-9pt}
\end{figure}

\vspace{-6pt}
In \autoref{fig:case-study}, we present a qualitative comparison. We observe that our model successfully captures the core strokes and structure of the unseen characters, including Oracle Bone Script examples. In contrast, baseline models often produce visually plausible but structurally incorrect shapes, highlighting the limitations of purely pixel-based understanding.

% Axis Ablation Table
\begin{wraptable}{r}{0.45\textwidth}
  \vspace{-36pt}
  \caption{Ablation results on the coordinate axis. The metric is the G$\uparrow$.}
  \label{tab:axis-ablation}
  \centering
  \resizebox{.45\textwidth}{!}{  
  \begin{tabular}{@{\hspace{0cm}}lc@{\hspace{0.2cm}}c@{\hspace{0.2cm}}c@{\hspace{0cm}}}
    \toprule
    \textbf{Model / Method} & \textbf{STD-zh} & \textbf{Stylistic-zh} & \textbf{OBS} \\
    \midrule
    \multicolumn{4}{c}{\cellcolor[HTML]{EFEFEF} \textit{Zero-shot Baselines}} \\
    Qwen2.5-VL-7B (w/o Axis) & 0.106 & 0.146 & 0.184 \\
    Qwen2.5-VL-7B (w/ Axis) & \textbf{0.157} & \textbf{0.177} & \textbf{0.182} \\
    \midrule
    GPT-4o (w/o Axis) & 0.347 & 0.361 & 0.381 \\
    GPT-4o (w/ Axis) & \textbf{0.431} & \textbf{0.467} & \textbf{0.409} \\
    \midrule
    \multicolumn{4}{c}{\cellcolor[HTML]{EFEFEF} \textit{Our Fine-tuned Model}} \\
    Ours (SFT w/o Axis) & 0.808 & 0.711 & 0.556 \\
    \textbf{Ours (SFT w/ Axis)} & \textbf{0.821} & \textbf{0.723} & \textbf{0.568} \\
    \bottomrule
  \end{tabular}
  }
\vspace{-18pt}
\end{wraptable}

% \vspace{-8pt}
\subsection{Ablation Studies}
% \vspace{-8pt}
To dissect our framework, we perform two key ablation studies in this section.
% Two ablation studies further dissect our framework.

% \vspace{-6pt}
\paragraph{The Critical Role of the Coordinate Axis.}
An ablation study (\autoref{tab:axis-ablation}) shows that the coordinate system is a crucial component. Removing the axes degrades performance not only for our fine-tuned model but also for the zero-shot baselines (Qwen2.5-VL and GPT-4o). This consistently positive effect suggests that the axes serve as a fundamental anchor for spacial grounding, enhancing the innate geometric understanding capabilities of VLMs even without task-specific training.

\begin{wraptable}{r}{0.45\textwidth}
  \vspace{-7pt}
  \caption{Ablation results comparing different methods training out model. Qwen2.5-VL-7B is employed for this study. The metric is the G$\uparrow$.}
  \label{tab:method-ablation}
  \centering
  \resizebox{.45\textwidth}{!}{
  \begin{tabular}{l@{\hspace{0.2cm}}c@{\hspace{0.2cm}}c@{\hspace{0.2cm}}c}
    \toprule
    \textbf{Model / Method} & \textbf{STD-zh} & \textbf{Stylistic-zh} & \textbf{OBS} \\
    \midrule
    {Ours (RL)} & {0.539} & {0.497} & {0.432} \\
    {Ours (Warmup + RL)} & {0.678} & {0.663} & {0.548} \\
    \textbf{Ours (SFT)} & \textbf{0.821} & \textbf{0.723} & \textbf{0.568} \\
    \bottomrule
  \end{tabular}
  }
  \vspace{-8pt}
\end{wraptable}

% \vspace{-6pt}
\paragraph{A Deeper Look into Reinforcement Learning.}\label{sec:rl}
We hypothesized that Reinforcement Learning (RL) would refine the SFT model's precision, but our results (\autoref{tab:method-ablation}) are more nuanced. While RL with an SFT warm-up improves over RL-only, it fails to surpass the performance of full SFT. This finding suggests a fundamental challenge in applying conventional RL algorithms to tasks involving fine-grained, continuous parameter generation. The action space for Bézier curve synthesis is vast, and the corresponding reward landscape is non-convex and sparse, where perturbations in a control point can cause discontinuous changes in the geometric score. This makes the credit assignment problem difficult for the exploration. In contrast, supervised learning provides a dense and globally coherent signal from ground-truth programs, which appears to be more effective in learning the complex structural dependencies required for this task. This highlights a key challenge for applying RL to structured visual generation and suggests that future investigations could benefit from exploring offline RL methodologies or developing more reward-shaping techniques tailored to geometric fidelity.

\vspace{-6pt}
\section{Conclusion}
\vspace{-6pt}
This paper demonstrates a Vision-Language Model's capacity to perform ``visual decompilation'' by translating raster images into a precise mathematical representation as executable geometric programs. The model's ability to reconstruct Oracle Bone Script, despite being trained solely on modern Chinese characters, provides evidence that it acquires an abstract and transferable geometric grammar rather than memorizing pixel configurations. This study validates a mathematically grounded programmatic approach to visual understanding, highlighting two key findings: the critical role of an explicit coordinate system for spatial grounding and the efficacy of supervised learning to generate structured outputs. Ultimately, this work contributes to the development of advanced AI systems with a structural and generalizable capability for visual interpretation.

While our approach demonstrates strong generalization across scripts and styles, several challenges remain for future work. First, a deeper mathematical analysis of the Bézier representation and its parameter sensitivity is needed to better understand the underlying geometry encoding. Second, current geometric reasoning still lacks rigorous theoretical justification, which could further strengthen the interpretability of the model. Finally, the inference latency and tokenization bottleneck introduced by long Bézier sequences highlight an efficiency limitation that calls for more compact and adaptive vector representations.

\section*{Acknowledgement}
This work is supported by the National Natural Science Foundation of China (No. 62276152, 62236011). We would like to thank Yaluo Liu for her assistance with the final revision of this paper.

% \clearpage
\thispagestyle{empty}

\bibliographystyle{plainnat}

\bibliography{references}

\appendix

\section{Bézier Curve Fundamentals}\label{sec:appendix_bezier}

A Bézier curve is a parametric curve fundamental to computer graphics, using Bernstein Polynomials as its basis. The curve, denoted as $c(t)$, is mathematically defined as a weighted sum of $n+1$ control points $\mathbf{b}_i$:
\begin{equation} \label{eq:bezier_appendix}
    c(t) = \sum_{i=0}^{n} \mathbf{b}_i B_{i,n}(t), \quad 0 \leq t \leq 1,
\end{equation}
where $n$ is the degree of the curve and $B_{i,n}(t)$ are the Bernstein basis polynomials. Unlike raster images composed of a static pixel grid, a Bézier curve provides a continuous, mathematical definition of a shape, determined entirely by its set of control points.

In our work, we leverage Bézier curves to model the individual strokes of a character. This approach is advantageous because it provides a compact, flexible, and scalable vector representation. These properties have made Bézier curves a powerful tool in various deep learning applications. For instance, BézierSketch utilize sequences of curves to produce scalable vector sketches, shifting the modeling paradigm from dense pixel grids to sparse control points~\citep{beziersketch}. More commonly, Bézier curves are used to parameterize and fit the external contours of objects. This is seen in domains such as real-time text spotting, where they define the boundaries of arbitrarily shaped text~\citep{Liu_2020_CVPR}, and medical image analysis, where they ensure smooth and continuous segmentation masks~\citep{life13030743}. In contrast to these approaches that focus on external boundaries, our work uses Bézier curves to reconstruct the internal topological structure of a character. By describing a character as a sequence of strokes, we transform the visual recognition problem into one of geometric program synthesis.

\section{Bézier Curve Extraction Details}\label{sec:python_libraries}

Our pipeline uses OpenCV~\citep{opencv_library} to load a grayscale raster and binarize it so strokes become foreground pixels. scikit-image’s~\citep{van2014scikit} skeletonize function collapses thick strokes to one-pixel centerlines while preserving topology. NetworkX~\citep{hagberg2008exploring} builds an 8-connected pixel graph (nodes are skeleton pixels, edges are neighborhood adjacencies). svgwrite exports each simplified segment as SVG paths. svgpathtools parses the resulting SVG and exposes Line, QuadraticBezier, and CubicBezier primitives so the pipeline can collect control points, merge aligned segments, and normalize coordinates.

\section{Experimental Details}\label{sec:appendix_exp}

\paragraph{Baselines} We compare our trained model against two powerful, zero-shot baselines: the base Qwen2.5-VL-7B model~\citep{bai2025qwen25vltechnicalreport} and GPT-4o~\citep{openai2024gpt4ocard}. For a robust and fair comparison, all baseline evaluations reported in our main results table are conducted using their stronger configuration, where the input image is enhanced with the same coordinate axis overlay provided to our models.

\paragraph{Implementation Details} All models are trained on 8 NVIDIA A800 GPUs using DeepSpeed ZeRO Stage 3 and BF16 precision. For our main SFT model, we train for 5 epochs using a learning rate of $1 \times 10^{-6}$ with a cosine scheduler and a warmup ratio of 0.01. We use a total batch size of 16. The vision transformer (ViT) backbone is kept frozen throughout training. For our comparative RL experiments, we initialize the model from a 5-epoch SFT warmup on $ \frac{1}{6} $ of the data, followed by 1 epoch of training with the GRPO algorithm on the remaining $ \frac{5}{6} $ of the data. The RL stage employs a distinct set of hyperparameters, including a global batch size of 48, a mini batch size of 24, rollout number per prompt of 6, and a generation temperature of 0.9, to explore the policy space.

\section{Detailed Evaluation Metrics}\label{sec:appendix_eval}

Our comprehensive scoring function, \textit{Geometric Score}, is designed to quantitatively evaluate the quality of a reconstruction by reflecting its geometric fidelity. The calculation involves a three-step process: calculating pairwise similarity matrices between strokes, finding the optimal global assignment for each metric, and normalizing the results.

\paragraph{Pairwise Stroke Similarity}
Given a generated sequence of Bézier curves and a ground truth sequence, we compute similarity matrices where each element $(i, j)$ represents a geometric similarity between the $i$-th ground truth stroke and the $j$-th generated stroke. For our analysis, we compute four distinct matrices: one for each sub-metric (Distance, Angle, Length), and a fourth composite matrix for the final \textit{Geometric Score}. The core components are calculated as follows:

\begin{enumerate}[leftmargin=*]
    \item \textbf{Sampling:} We uniformly sample 10 points and their corresponding tangent vectors from both Bézier curves.
    \item \textbf{Reward Calculation:} We compute three distinct geometric rewards, all normalized to $[0, 1]$, which serve as the values in the sub-metric similarity matrices:
    \begin{itemize}
        \item \textbf{Distance Reward:} Calculated as $1 / (1 + \bar{d})$, where $\bar{d}$ is the mean Euclidean distance between corresponding sampled points.
        \item \textbf{Length Reward:} Calculated as $\min(L_1, L_2) / \max(L_1, L_2)$, rewarding similarity in the total arc length of the curves.
        \item \textbf{Angle Reward:} The mean cosine similarity between corresponding tangent vectors, mapped from $[-1, 1]$ to $[0, 1]$. This measures directional alignment.
    \end{itemize}
    \item \textbf{Weighted Combination for Geometric Score:} For the composite similarity matrix used to calculate the final \textit{Geometric Score}, the three rewards are combined using predefined weights: $0.6 \times \text{Distance} + 0.2 \times \text{Length} + 0.2 \times \text{Angle}$. We also account for stroke directionality by calculating this composite score for both the original and reversed generated stroke, taking the maximum of the two.
\end{enumerate}

\paragraph{Optimal Stroke Matching}
With the pairwise similarity matrices constructed, we treat the problem as a maximum weight bipartite matching task. This step is applied to each of the four matrices independently. We use the Hungarian algorithm~\citep{https://doi.org/10.1002/nav.3800020109} to find the optimal one-to-one assignment of generated strokes to ground truth strokes that maximizes the total similarity for that specific metric. This approach ensures that each reported score (Geometric Score, Distance, Angle, Length) is based on a globally optimal matching for that specific criterion. It is important to note that the final \textit{Geometric Score} is derived from the matching on the composite matrix and is therefore not a simple weighted average of the three final sub-scores.

\paragraph{Final Score Normalization}
The sum of similarities from the optimal matching for each metric is normalized by the maximum number of strokes between the ground truth and the generated sequence. This yields a base score in the range $[0, 1]$. To enhance the signal for reinforcement learning, the score is then passed through a normalized sigmoid function centered at a threshold of 0.8, which amplifies differences in high-quality reconstructions.

\section{Human Evaluation Details}\label{sec:appendix_human_eval}

To validate that our proposed \textit{Geometric Score} aligns with human perception of reconstruction quality, we conducted a formal human evaluation. This evaluation was designed to compare the qualitative performance of our best SFT model against the strong GPT-4o baseline.

\paragraph{Evaluation Setup}
We randomly sampled a total of 150 test cases, comprising 50 cases from each of our three distinct evaluation datasets: \textbf{Chinese STD}, \textbf{Chinese Stylistic}, and \textbf{OBS}. The evaluation was performed by five human experts, defined as individuals with proficiency in both Chinese calligraphy and digital vector graphics. The detailed breakdown of choices from each expert is shown in \autoref{tab:human-eval-breakdown}.

\begin{table}[h]
  \caption{Detailed breakdown of scores from the five human evaluators. ``W'' denotes cases where our model was preferred, ``T'' for ties, and ``L'' where the baseline was preferred.}
  \label{tab:human-eval-breakdown}
  \centering
  \begin{tabular}{lcccc}
    \toprule
    \textbf{Evaluator} & \textbf{W} & \textbf{T} & \textbf{L} & \textbf{Win Rate} \\
    \midrule
    Annotator 1 & 142 & 7  & 1 & 97.00\% \\
    Annotator 2 & 139 & 8  & 3 & 95.33\% \\
    Annotator 3 & 128 & 21 & 1 & 92.33\% \\
    Annotator 4 & 113 & 32 & 5 & 86.00\% \\
    Annotator 5 & 114 & 34 & 2 & 87.33\% \\
    \bottomrule
  \end{tabular}
\end{table}

\paragraph{Evaluation Protocol}
The evaluation followed a blind, side-by-side comparison protocol. For each of the 150 cases, the experts were presented with the ground-truth character image. Alongside it, the reconstructions from our SFT model and GPT-4o were displayed in a randomized order to prevent positional bias. The experts were not informed which model generated which image.

Following a standard methodology for evaluating generative models~\citep{NEURIPS2023_91f18a12}, experts were asked to make a three-way choice for each pair of reconstructions:
\begin{itemize}[leftmargin=*]
    \item \textbf{Model A is Better}: If one reconstruction was clearly superior.
    \item \textbf{Model B is Better}: If the other reconstruction was clearly superior.
    \item \textbf{Tied}: If both reconstructions were of comparable quality (either equally good or equally bad).
\end{itemize}
The judgment criteria provided to the experts focused on (1) preservation of the character's topological structure, and (2) overall visual similarity to the source image.

\paragraph{Win Rate Calculation}
To aggregate the results, we calculated the overall win rate for our model against the baseline. The win rate is computed using a standard formula that gives partial credit for ties, providing a more nuanced performance measure than a simple win-loss metric. The formula is as follows:
\begin{equation}
    \text{Win Rate} = \frac{\#(\text{Our Model Wins}) + 0.5 \times \#(\text{Ties})}{\text{Total Number of Comparisons}}
\end{equation}
The final win rate, aggregated across all 150 samples and 6 experts, is reported in the main body of the paper.

\section{Related Work}

\paragraph{Visual Program Synthesis.} This field aims to translate visual inputs into executable programs, with diverse approaches tackling the challenge. ~\citet{10.5555/3327345.3327505} established the idea of factorizing the problem into a neural perception stage to identify drawing primitives and a symbolic synthesis stage to assemble them into a coherent program. More recent methods include reinforced self-training to overcome the need for expert-annotated datasets~\citep{khan2024self}, while other novel paradigms explore neuro-symbolic languages~\citep{pldi2023imageeye} or diffusion models that operate directly on program syntax trees~\citep{kapur2024diffusion}. In contrast, our work introduces a fine-grained ``visual decompiler'' task, training a model to translate images directly into a low-level geometric program of Bézier curves, emphasizing direct, structured generation over interpretative reasoning.

\paragraph{Generative Models for Vector Graphics.} Generating vector graphics is a significant challenge, with key approaches including dual-modality learning~\citep{wang2021deepvecfont} and cascaded diffusion models like VecFusion~\citep{thamizharasan2024vecfusion}, which often rely on an intermediate raster stage to guide the final vector output. Other models like StrokeNUWA generate Scalable Vector Graphics (SVG) code directly, using ``stroke tokens'' that are inherently compatible with Large Language Models (LLMs) and mimic the sequential process of a human artist~\citep{10.5555/3692070.3694021}. Hybrid frameworks such as Chat2SVG combine the semantic reasoning of LLMs to generate a basic template with the refinement power of image diffusion models to add geometric detail~\citep{Wu_2025_CVPR}. Our framework differs by focusing on direct, end-to-end programmatic reconstruction from an image, bypassing creative generation and intermediate pixel-based stages to learn a canonical geometric grammar for a given character.

\paragraph{Vision-language Models for Text in Images.}
Historically, understanding text in images for tasks like Visual Question Answering (VQA) relied on multi-stage pipelines that sequentially performed detection, Optical Character Recognition (OCR), and reasoning. Modern Vision-language Models (VLMs) aim to replace this complex, error-prone process with a single end-to-end model~\citep{lamm2024visuallanguagemodelsreplace}. However, this unified approach often struggles with fine-grained or dense text, where performance can be unreliable for tasks beyond simple transcription~\citep{chen2025oceanocrgeneralocrapplication, lamm2024visuallanguagemodelsreplace}. This has led to explorations into integrating specialized OCR encoders to improve detail preservation~\citep{Nacson_2025_CVPR}. A deeper level of understanding moves beyond mere transcription to parse the internal structure of characters; for complex logograms like Chinese, for instance, this involves recognizing constituent radicals and stroke arrangements~\citep{wu-etal-2025-impact}. Our work advances this trajectory with a fundamentally different paradigm. Instead of asking the model to recognize or classify visual text, we frame the task as one of programmatic reconstruction, compelling the model to generate an executable geometric program. This shifts the goal from semantic identification to learning an underlying, generative grammar of form.

\paragraph{Historical Script Understanding and Generalization.} 
Recognizing or understanding historical scripts is a difficult task due to data scarcity, degradation, and stylistic variation. This is particularly true for logographic systems like Oracle Bone Script (OBS), which have vast and complex character sets~\citep{diao2025ancientscriptimagerecognition}. Traditional approaches often rely on script-specific, multi-stage pipelines for classification~\citep{ijcai2020p779} or general zero-shot recognizers like CLIP~\citep{radford2021learning}. 
More recently, advanced frameworks have begun to leverage Vision-Language Models for this challenge. For instance, V-Oracle frames the task as a visual question-answering problem to interpret scripts through a multi-step reasoning chain~\citep{qiao-etal-2025-v}, while OracleFusion generates ``semantically enriched vector fonts'' for OBS, extending from analysis to creation~\citep{li2025oraclefusionassistingdeciphermentoracle}. 
We build upon this direction but propose a novel paradigm of generalization through generation. By training on modern characters, our model learns a transferable, ``universal geometric grammar.'' This allows it to demonstrate a deep, structural understanding by programmatically reconstructing an unseen script, rather than merely classifying or interpreting it, pushing the boundaries of cross-script generalization.

\end{document}